\PassOptionsToPackage{table,dvipsnames}{xcolor}
\documentclass[sigconf]{acmart}

\usepackage[table,dvipsnames]{xcolor} %
\usepackage[misc,geometry]{ifsym}
\usepackage{microtype}
\usepackage{graphicx}
\usepackage{subfigure}
\usepackage{booktabs} %
\usepackage{natbib}  %
\usepackage{caption} %
\usepackage{amsmath}
\usepackage{float}
\usepackage{algorithm}
\usepackage{algorithmic}

\newcommand\extrafootertext[1]{%
    \bgroup
    \renewcommand\thefootnote{\fnsymbol{footnote}}%
    \renewcommand\thempfootnote{\fnsymbol{mpfootnote}}%
    \footnotetext[0]{
    \vspace{-0.7cm}
    \par#1}%
    \egroup
}

\AtBeginDocument{%
  }

\renewcommand\footnotetextcopyrightpermission[1]{} %

\begin{document}
\fancyhf{} %
\pagestyle{empty} %

\title{KMM: Key Frame Mask Mamba for Extended Motion Generation}

\author{Zeyu Zhang$^{12*\dag}$\quad Hang Gao$^{3*}$\quad Akide Liu$^3$\quad Qi Chen$^4$\quad Feng Chen$^4$\quad Yiran Wang$^5$\\
Danning Li$^6$\quad Rui Zhao$^7$\quad Zhenming Li$^8$\quad Zhongwen Zhou$^8$\quad Hao Tang$^{1\text{\Letter}}$\quad Bohan Zhuang$^9$\\
\vspace{0.2cm}
$^1$PKU\quad $^2$ANU\quad $^3$Monash\quad $^4$UoA AIML\quad $^5$USYD ACFR\quad $^6$McGill\quad $^7$JD.com\quad $^8$AI Geeks\quad $^9$ZJU\\
\vspace{0.2cm}
{\bfseries\fontsize{11}{4}\selectfont\url{https://steve-zeyu-zhang.github.io/KMM}}
}

\begin{abstract}
\extrafootertext{\noindent\vspace{0.5em}
\\$^{*}$Equal contribution. $^{\dag}$Work done while being a visiting student researcher at Peking University. $^{\text{\Letter}}$Corresponding author: \href{mailto:bjdxtanghao@gmail.com}{bjdxtanghao@gmail.com}}
Human motion generation is an advanced area of research in generative computer vision, driven by its promising applications in video creation, game development, and robotic manipulation. As an effective solution for modeling long and complex motion sequences, the recent Mamba architecture has demonstrated significant potential, yet two major challenges remain:
Firstly, generating long motion sequences poses a challenge for Mamba, as its implicit memory architecture suffers from capacity limitations, leading to underperform over extended motions. Secondly, compared to Transformers, Mamba struggles with effectively aligning motions with textual queries, often resulting in errors such as confusing directions (e.g. left or right) or failing to capture details from longer text descriptions.
To address these challenges, our paper presents three key contributions: Firstly, we introduce \textbf{KMM}, a novel architecture featuring \textbf{K}ey frame \textbf{M}asking \textbf{M}odeling, designed to enhance Mamba's focus on key actions in motion segments. 
This approach enhances the motion representation of Mamba and ensures that the memory of the hidden state focuses on the key frames.
Additionally, we designed a fine-grained text-motion alignment mechanism, leveraging frame-level annotation to bring pairwise text and motion features closer in the representation space.
Finally, we conducted extensive experiments on multiple datasets, achieving state-of-the-art performance with a reduction of more than $\mathbf{0.24}$ in FID while using $\mathbf{55\%}$ fewer parameters and reducing GFLOPs by $\mathbf{70\%}$ compared to previous state-of-the-art methods. This demonstrates that our method achieves superior performance with greater efficiency.
\end{abstract}


\begin{teaserfigure}
  \includegraphics[width=\textwidth]{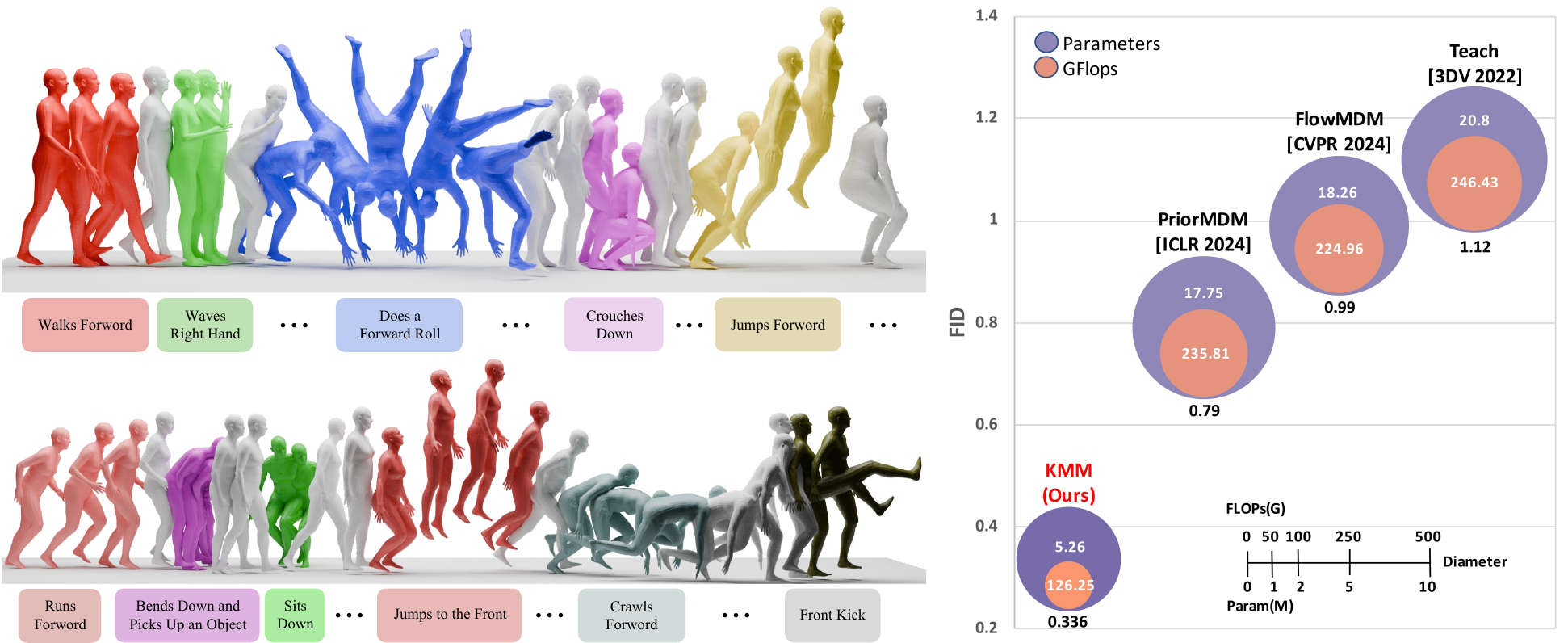}
  \caption{The figure on the left illustrates the exceptional capability of the proposed KMM in generating continuous and diverse human motions based on extended text prompts across various durations. The figure on the right highlights that our method significantly outperforms the previous state-of-the-art in quantitative evaluations while utilizing substantially fewer FLOPs.}
  \label{fig:main}
\end{teaserfigure}

\maketitle

\section{Introduction}
\label{submission}

Text-to-motion (T2M) generation \cite{guo2022generating} involves creating realistic 3D human movements from text descriptions, with promising applications in game development, video creation, and digital humans. Previous generation methods that leverage VAE \cite{guo2022generating,petrovich2022temos,zhong2023attt2m,tevet2022motionclip}, GAN \cite{lin2018human,harvey2020robust,barsoum2018hp}, autoregressive \cite{jiang2024motiongpt,pinyoanuntapong2024mmm,guo2024momask}, and diffusion-based \cite{zhang2024motiondiffuse,shafir2023human,chen2023executing,zhang2023remodiffuse} approaches have achieved unprecedented success in downstream tasks \cite{raab2023single,zhang2024motionavatar,xiao2023unified}. However, long motion generation is still not well addressed by these conventional methods, since it involves generating coherent, complex motion sequences conditioned on rich, descriptive text prompts as Figure \ref{fig:main}. Recently, Mamba \cite{gu2023mamba} has shown promising potential for efficient long-context modeling, thanks to its recurrent architecture and linear scaling with sequence length \cite{dao2024transformers}. Moreover, it has already achieved encouraging results in human motion grounding \cite{wang2024text} and generation \cite{zhang2025motion,zhang2024infinimotion}. However, leveraging Mamba for long motion generation presents two significant challenges: 

(1) First, the memory matrix in Mamba’s hidden states has limited capacity for retaining implicit memory, which is insufficient for modeling complex and long motions compared to Transformers \cite{zhang2025motion}, leading to underperformance when generating entire long motion sequences. For example, when testing on complex and lengthy text prompts, models often fail to generate sufficient motions corresponding to the instructions or omit latter part of the text.

\begin{figure*}[t]
    \centering
    \includegraphics[width=\linewidth]{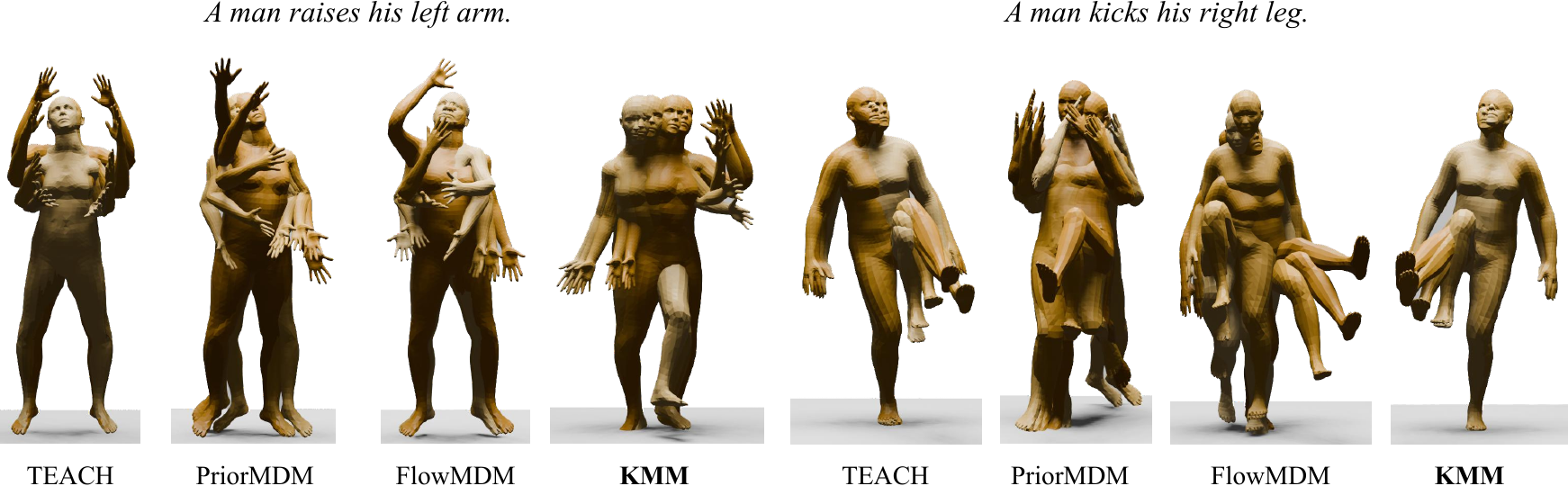}
    \caption{The figure illustrates that previous extended motion generation methods often struggle with directional instructions, leading to incorrect motions. In contrast, our proposed KMM, with enhanced text-motion alignment, effectively improves the model's understanding of text queries, resulting in more accurate motion generation.}
    \label{fig:compare}
\end{figure*}

(2) Second, Mamba intrinsically struggles with multimodal fusion due to its sequential architecture \cite{wang2024text,zhang2025motion}, which is less effective than that of Transformers \cite{dong2024fusion, xie2024fusionmamba}. This results in poor alignment between text and motion, ultimately decreasing generation performance. One typical scenario of text-motion misalignment is the misunderstanding of directional instructions. For instance, when tested on queries containing directions such as left and right, models often generate incorrect or opposite directional motions, as illustrated in Figure~\ref{fig:compare}.

To address the first challenge, we design a key frame masking strategy that allows the model to focus on learning the key actions within a long motion sequence, fully utilizing the limited implicit memory of Mamba. 
Our key frame masking computes local density and pairwise distances to selectively mask high-density motion embeddings in the latent space. This approach is more effective than other masked motion modeling approaches \cite{pinyoanuntapong2024mmm,guo2024momask,pinyoanuntapong2024bamm} because it helps the model focus on learning key frames. Although key frame learning in motion generation has been explored by works like Diverse Dance \cite{pan2021diverse} and KeyMotion \cite{geng2024text}, our method fundamentally differs from these previous approaches in selection and learning of key frames. Diverse Dance uses key frames as conditions to generate motion sequences around them. Similarly, KeyMotion treats key frames as anchors, generating key frames first and then performing motion infilling to complete the sequence. In contrast, our method introduces a novel key frame selection technique based on local density, selecting high-density motion tokens as key frames. Instead of treating these key frames as conditions or anchors, we mask them out to enhance learning of motion representation.

To address the second challenge, we design a contrastive loss between fine-grained texts and motion segments to enhance text-motion alignment.  Although there have been attempts to address the multimodal fusion problem in Mamba for human motion modeling, such as using a transformer mixer \cite{zhang2025motion} or modifying selective scan \cite{wang2024text}, the results remain unsatisfactory. There are still misalignments between text descriptions and motion, especially when dealing with directions such as left and right or when the text queries are complex. Moreover, the misalignment between text and motion is not unique to the Mamba architecture, it is a common issue that also affects other Transformer-based diffusion and autoregressive methods, as illustrated in Figure \ref{fig:compare}. Despite variety on architecture, existing methods share a common approach, they use a frozen CLIP text encoder to learn a shared latent space for text and motion. This inspired us to improve text-motion alignment by designing a robust contrastive learning paradigm that consistently learns the correspondence between motion and text, rather than relying on a frozen CLIP encoder.

To overcome these challenges, our paper presents three key contributions:

\begin{itemize}
    \item Firstly, to address the memory limitations of Mamba's hidden state, we introduce \textbf{K}ey frame \textbf{M}asking \textbf{M}odeling (\textbf{KMM}), a novel approach that selects key frames based on local density and pairwise distance. This method allows the model to focus on learning from masked key frames, which is more effective for the implicit memory architecture of Mamba than random masking. This advancement represents a pioneering method that customizes frame-level masking in the Mamba model within the latent space.
    \item Additionally, to address the issue of poor text-motion alignment in the Mamba architecture caused by ineffective multimodal fusion, we proposed a novel method that leverages contrastive learning. Instead of relying on a fixed CLIP text encoder, our approach dynamically learns text encodings, enabling the generation of more accurate motions by encoding text queries with better alignment.
    \item Lastly, we conducted extensive experiments across multiple datasets, achieving state-of-the-art performance with an FID reduction of over $\mathbf{0.24}$ while utilizing $\mathbf{55\%}$ fewer parameters and lowering GFLOPs by $\mathbf{70\%}$ compared to previous state-of-the-art methods, as shown in Figure \ref{fig:main}. These results highlight the superior performance and improved efficiency of our approach.
\end{itemize}

\section{Related Works}

\paragraph{Text-to-Motion Generation.}
Autoencoders have been essential to motion generation. For example, JL2P \cite{ahuja2019language2pose} employs RNN-based autoencoders \cite{hopfield1982neural} for a unified language-pose representation, albeit with a strict one-to-one mapping. MotionCLIP \cite{tevet2022motionclip} utilizes Transformer-based autoencoders \cite{vaswani2017attention} to reconstruct motion aligned with text in the CLIP \cite{radford2021learning} space. Transformer-based VAEs \cite{kingma2014auto} in TEMOS \cite{petrovich2022temos} and T2M \cite{guo2022generating} generate latent distribution parameters, while AttT2M \cite{zhong2023attt2m} and TM2D \cite{gong2023tm2d} integrate body-part spatio-temporal encoding into VQ-VAE \cite{van2017neural} for richer discrete representations. 

Diffusion models \cite{sohl2015deep, ho2020denoising, dhariwal2021diffusion, rombach2022high} have been adapted for motion generation: MotionDiffuse \cite{zhang2024motiondiffuse} introduces a probabilistic, multi-level diffusion framework; MDM \cite{tevet2022human} employs a classifier-free Transformer-based model predicting samples rather than noise; and MLD \cite{chen2023executing} applies diffusion in the latent space. Recently, Motion Mamba \cite{zhang2025motion} exploits hierarchical SSMs for efficient long-sequence generation. Additionally, 

Transformer-based approaches such as MotionGPT \cite{jiang2024motiongpt} treat motion as a “foreign language,” while masked modeling methods in MMM \cite{pinyoanuntapong2024mmm} and MoMask \cite{guo2024momask}, alongside the bidirectional autoregression in BAMM \cite{pinyoanuntapong2024bamm}, further enhance motion generation.

\paragraph{Extended Motion Generation.}
Recent studies focus on producing long, coherent motion sequences. MultiAct \cite{lee2023multiact} pioneers long-term 3D human motion generation from multiple action labels, and TEACH \cite{athanasiou2022teach} introduces a temporal action composition framework for fine-grained control. In the diffusion realm, PriorMDM \cite{shafir2023human} employs generative priors while DiffCollage \cite{zhang2023diffcollage} utilizes parallel generation for large-scale content. Transformer-based models, exemplified by T2LM \cite{lee2024t2lm} and InfiniMotion \cite{zhang2024infinimotion}, extend synthesis to complex narratives by enhancing memory capacity with the Mamba architecture. Moreover, FlowMDM \cite{barquero2024seamless} leverages blended positional encodings, PCMDM \cite{yang2023synthesizing} introduces coherent sampling techniques, and STMC \cite{petrovich2024multi} offers multi-track timeline control, collectively advancing the coherence and diversity of extended motion sequences.

\begin{figure*}[t]
    \centering
    \includegraphics[width=\linewidth]{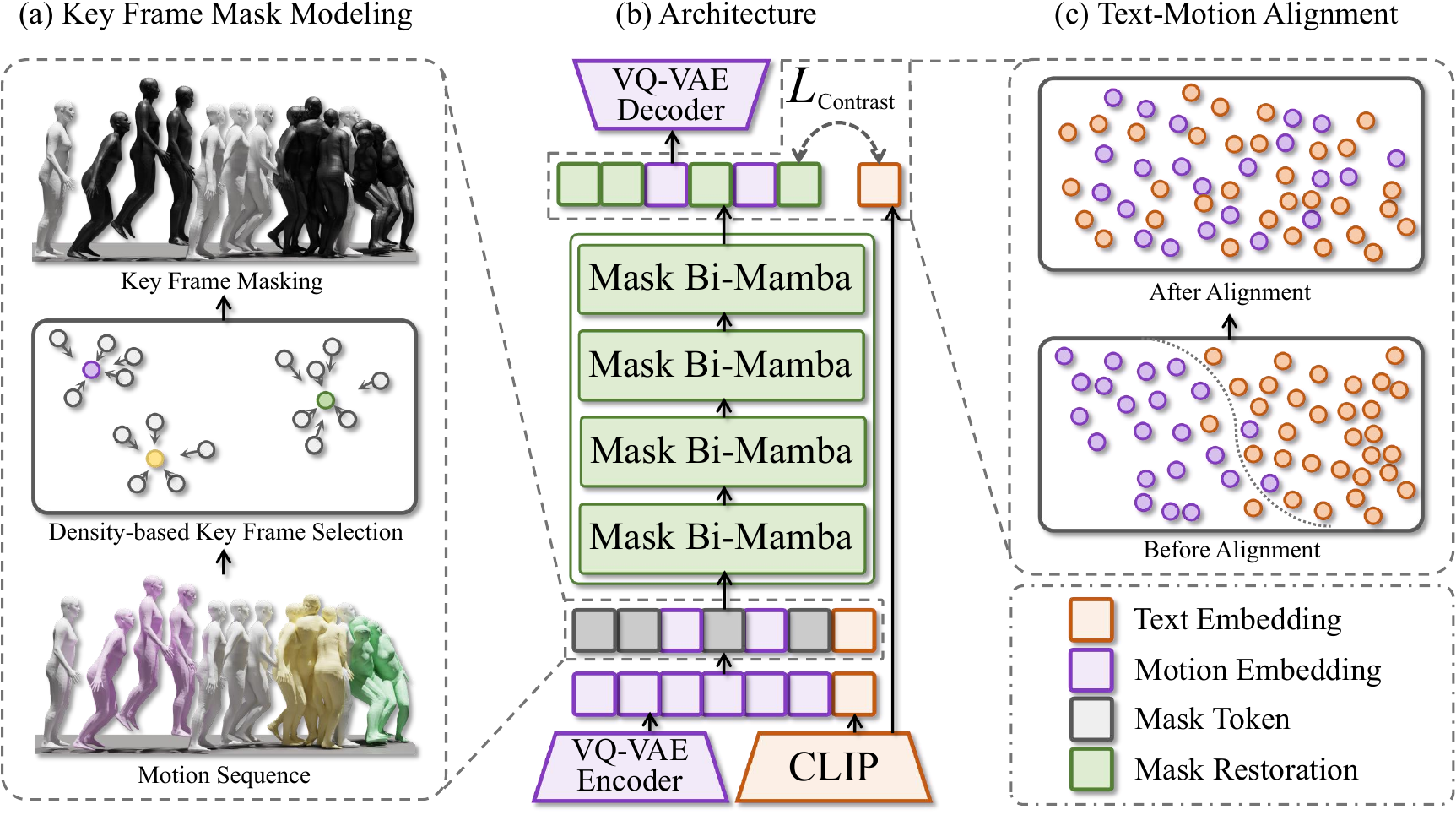}
    \caption{The figure demonstrates our novel method from three different perspectives: (a) illustrates the key frame masking strategy based on local density and minimum distance to higher density calculation. (b) showcases the overall architecture of the masked bidirectional Mamba. (c) demonstrates the text-to-motion alignment, highlighting the process before and after alignment.}
    \label{fig:arch}
\end{figure*}

\section{Methodology}

\subsection{Overview}

The overall architecture is an autoregressive model for long-motion generation. During training, the motion sequence is first compressed into a latent space using VQ-VAE with a codebook, followed by token masking with key-frame mask modeling. The motion tokens are then concatenated with the text embedding (CLIP token) and processed by a four-layer Mask Bi-Mamba for masked restoration. Meanwhile, frame-level text-motion alignment is performed to enhance the model's ability to understand and capture the text prompt, as shown in Figure \ref{fig:arch} and Algorithm \ref{alg:KMM}.

\subsection{Key Frame Mask Modeling}

Our proposed key frame masking model introduces a novel density-based key frame selection and masking strategy. First, we calculate the local density of each temporal token, then consecutively find the minimum distance to higher density. This process allows us to identify the tokens with the highest density as the key frame and mask them out.

\paragraph{Local Density Calculation.} Let $\mathbf{X} \in \mathbb{R}^{n \times l}$ denotes the motion embedding in the latent space, where $n$ refers to the number of token in temporal dimension, and $l$ refers to the spatial dimension.
\begin{equation}
    \mathbf{X} = (\mathbf{x}_1, \mathbf{x}_2,..., \mathbf{x}_n), \quad \mathbf{x}_i \in \mathbb{R}^l
\end{equation}

We first compute the pairwise Euclidean distance matrix $\mathbf{D} \in \mathbb{R}^{n \times n}$.
\begin{equation}
    \mathbf{D}_{i,j} = ||\mathbf{x}_i - \mathbf{x}_j||_2 = \sqrt{\sum_{k=1}^{l}{(\mathbf{x}_{i,k} - \mathbf{x}_{j,k})^2}},
\end{equation}
where $\mathbf{x}_i$ and $\mathbf{x}_j$ are the $i$-th and $j$-th rows of $\mathbf{X}$, $\mathbf{x}_{i,k}$ and $\mathbf{x}_{j,k}$ are the $k$-th element of $\mathbf{x}_i$ and $\mathbf{x}_j$. 

Then the local density $\mathbf{d} \in \mathbb{R}^{n}$ could be calculated as
\begin{equation}
    \mathbf{d}_i = \sum_j{\exp{(-\mathbf{D}^2_{i,j})}},
\end{equation}
which represents the sum of Gaussian kernel values centered as each latent vector $\mathbf{x}_i$, where the kernel bandwidth is determined by the squared distance $\mathbf{D}^2_{i,j}$.

Hence, the local density for the $i$-th token can be summarized by
\begin{equation}
    \mathbf{d}_{i} = \sum_j \exp{(-||\mathbf{x}_{i} - \mathbf{x}_{j}||^2_2)},
\end{equation}
where $\mathbf{x}_{i}$ is the latent vector for the $i$-th token.

\paragraph{Minimum Distance to Higher Density.} We expand the local density $\mathbf{d}$ into two intermediate matrices $\mathbf{d}_\text{col}$ $\in \mathbb{R}^{1 \times n}$ and $\mathbf{d}_\text{row}$ $\in \mathbb{R}^{n \times 1}$ for broadcasting, ensuring that each column and row is a duplicate of the local density $\mathbf{d}$.

We then create a boolean mask matrix $\mathbf{M} \in \{0,1\}^{n \times n}$. Please note that this masking is intended to find the minimum distance to higher density, which is a different concept from masking frames.
\begin{equation}
\mathbf{M}_{i,j} =
    \begin{cases}
    1, \quad \text{if } \mathbf{d}_{\text{col},i} < \mathbf{d}_{\text{row},j}\\
    0, \quad \text{otherwise}
    \end{cases}
\end{equation}

This means that ${M}_{i,j}$ is $1$ (True) only if the local density of the $i$-th token is less than the local density of the $j$-th token.

We then apply the mask to distance matrix $\mathbf{D}$ in-place
\begin{equation}
\mathbf{D}_{i,j} =
    \begin{cases}
    \mathbf{D}_{i,j}, \quad \text{if } {M}_{i,j} = 1\\
    \infty,  \quad \text{if } {M}_{i,j} = 0
    \end{cases}
\end{equation}

This effectively sets all distances to infinity where the mask is 0 (False), meaning we discard distances from tokens to other tokens with lower or equal density.

The masking operation ensures that for each token $i$, we only consider distances to other tokens $j$ that have a strictly higher local density

\begin{equation}
    \mathbf{D}_\text{masked} = \mathbf{D} \odot \mathbf{M} + (\mathbf{1} - \mathbf{M}) \odot \infty,
\end{equation}
where $\odot$ is the element-wise (Hadamard) product and $\mathbf{1}$ is a matrix of all ones. This prepares the distance matrix for the subsequent step of finding the minimum distance to a higher-density token.

This can give us the masked distance matrix $\mathbf{D}_\text{masked} \in \mathbb{R}^{n \times n}$, where distances to lower or equal density tokens have been set to infinity. 

For each row $i$ (corresponding to each token), we find the minimum distance $\mathbf{S}$ along the columns of $\mathbf{D}_\text{masked}$:
\begin{equation}
    \mathbf{S}_i = \min_j{\mathbf{D}_{\text{masked},i,j}}.
\end{equation}
Due to the masking, this minimum value will be either:
\begin{itemize}
    \item The actual minimum Euclidean distance to a token with strictly higher density, if such a token exists.
    \item Infinity, if no token with higher density exists.
\end{itemize}

The resulting minimum distances are collected in $\mathbf{S} \in \mathbb{R}^{n}$, which represents the distance to a higher density for all frames. Hence, the minimum distance to higher density, denoted as $\mathbf{S}_i$ for the $i$-th token, is calculated as
\begin{equation}
    \mathbf{S}_i = \min_{j:\mathbf{d}_j>\mathbf{d}_i}{||\mathbf{x}_i - \mathbf{x}_j||_2}.
\end{equation}

\paragraph{Key Frame Masking.} After calculating the local density and the minimum distance to higher density, we can determine the density parameter for all temporal tokens, denoted as $\mathbf{\Gamma} \in \mathbb{R}^{n}$.
\begin{equation}
    \mathbf{\Gamma} = \mathbf{d} \odot \mathbf{S}, \Gamma_i = \mathbf{d}_i \cdot \mathbf{S}_i,
\end{equation}
where $\Gamma_i$ is the density parameter for the $i$-th token, $\mathbf{d}_i$ is the local density for the \textbf{i}-th token, and $\mathbf{s}_i$ is the distance to a higher density for the $i$-th token.

Hence, based on the density parameter \(\mathbf{\Gamma}\), we can select the temporal tokens with the highest density as the key frames in the motion latent space.
\begin{equation}
    \mathbf{K} = \underset{i}{\operatorname{argmax}} : \mathbf{\Gamma}_i,
\end{equation}
where $ \mathbf{K} $ is the index of the selected key frames, and $ \underset{i}{\operatorname{argmax}} \: \mathbf{\Gamma}_i $ represents the index corresponding to the maximum value in the $\mathbf{\Gamma}$ matrix.

After obtaining the key frame index \(\mathbf{K}\), we can perform a unidirectional mask along with the padding mask on Mamba's sequential architecture.

\begin{figure*}[t]
    \centering
    \includegraphics[width=\linewidth]{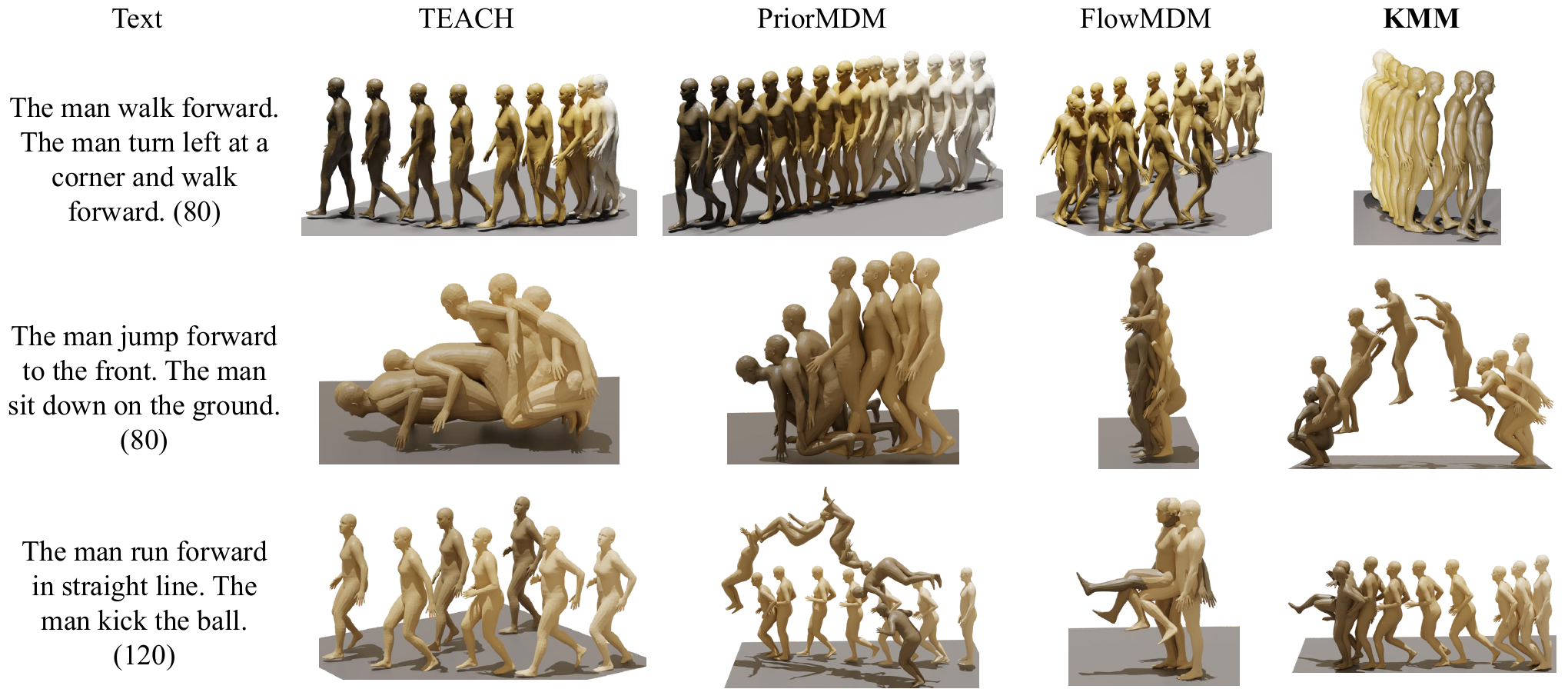}
    \caption{The figure demonstrates a qualitative comparison between the previous state-of-the-art method in extended motion generation and our KMM. The qualitative results show that our method significantly outperforms others in handling complex text queries and generating more accurate corresponding motions.}
    \label{fig:qualitative}
\end{figure*}

\begin{algorithm}[t]
\caption{KMM: Key Frame Mask Mamba}
\label{alg:KMM}
\centering
\scalebox{1}{
\begin{minipage}{\columnwidth}
\begin{algorithmic}[1]
\REQUIRE Motion embedding matrix $\mathbf{X} \in \mathbb{R}^{n \times l}$ ($\mathbf{x}_i \in \mathbb{R}^{l}$ for $i=1,\dots,n$); Text embeddings $\mathbf{T}$ from CLIP, learnable temperature $\tau$, and loss coefficient $\lambda$
\ENSURE Extended motion sequence aligned with text prompt

\STATE \textbf{// Compute Pairwise Euclidean Distance Matrix}
\[
\mathbf{D}_{i,j} = \|\mathbf{x}_i - \mathbf{x}_j\|_2, \quad \forall\, i,j=1,\dots,n.
\]

\STATE \textbf{// Local Density Calculation}
\[
\mathbf{d}_i = \sum_{j=1}^{n} \exp\left(-\mathbf{D}_{i,j}^2\right), \quad \forall\, i.
\]

\STATE \textbf{// Mask and Find Minimum Distance to Higher Density}
\FOR{$i=1$ \TO $n$}
  \STATE \[
  \mathbf{S}_i = \min_{j:\mathbf{d}_j>\mathbf{d}_i}\left\{ \mathbf{D}_{i,j} \ \text{if}\ \mathbf{d}_j>\mathbf{d}_i,\; \infty \ \text{otherwise} \right\}.
  \]
\ENDFOR

\STATE \textbf{// Key Frame Selection via Density Parameter}
\[
\Gamma_i = \mathbf{d}_i \cdot \mathbf{S}_i,\quad \mathbf{K} = \operatorname*{arg\,max}_{i} \Gamma_i.
\]

\STATE \textbf{// Text-Motion Alignment using Contrastive Loss}
For each sample pair $(i,j)$:
\[
\text{sim}_{ij} = \frac{\mathbf{T}_i^\top \mathbf{M}_j}{\tau}.
\]
Define labels $\mathbf{y} = [0,1,\ldots,b-1]$ and compute
\[
\mathcal{L}_{\text{contrast}} = \lambda \Big( \text{CE}(\text{sim}, \mathbf{y}) + \text{CE}(\text{sim}^\top, \mathbf{y}) \Big).
\]

\STATE \textbf{// Motion Generation}
Concatenate masked motion tokens and text embeddings, and process via a four-layer Mask Bi-Mamba network.

\RETURN Generated motion sequence.
\end{algorithmic}
\end{minipage}
}
\end{algorithm}

\subsection{Text-Motion Alignment}

Text-to-motion alignment remains a significant challenge in human motion generation tasks. This challenge arises because generation models, whether based on transformers or diffusion approaches, struggle to effectively understand the text features embedded by the CLIP encoder. This results in a misalignment between the text and motion modalities.
From a latent space perspective, motion generation models operate within two distinct latent spaces: the text features encoded by CLIP and the motion features generated by the motion model. The substantial gap between these two modalities represents a core challenge.
Most previous works leverage CLIP as a semantically rich text encoder, keeping it frozen while injecting text embeddings extracted from it into the generation model. In the context of multi-modal fusion, two latent spaces, \(z_1\) and \(z_2\), are typically aligned using an alignment mechanism \(f_{align}\). In our case, \(z_1\) and \(z_2\) correspond to the text latent space \(z_{\text{text}}\) and the motion latent space \(z_{\text{motion}}\), respectively.
In the common practice of motion generation tasks, the CLIP text encoder is frozen, and no explicit alignment mechanism is employed. Consequently, the generation model is implicitly required to learn the alignment between these modalities. However, since the generation model is not specifically designed to address the significant gap between the text and motion modalities, this often leads to misalignment.
To address this issue, we propose leveraging a contrastive learning objective to reduce the distance between these two latent spaces. This approach aims to decrease the learning difficulty and enhance the model's overall multi-modal capabilities and performance. To be more specific, our text-motion alignment can be described as follows:

Let \( \mathbf{T}_i \) be the text latents for the \( i \)-th sample, and \( \mathbf{M}_j \) be the motion latents for the \( j \)-th sample. The similarity between text latents \( \mathbf{T}_i \) and motion latents \( \mathbf{M}_j \) is calculated as:
\begin{equation}
   \text{sim}_{ij} = \mathbf{T}_i^\top \mathbf{M}_j.
\end{equation}

Then, the similarity is scaled by a learnable temperature parameter $\tau$:
\begin{equation}
   \text{sim}_{ij} = \frac{\mathbf{T}_i^\top \mathbf{M}_j}{\tau}.
\end{equation}

Furthermore, we define the contrastive labels as \( \mathbf{y} = [0, 1, 2, \dots, b-1] \). The contrastive loss for text and motion embedding can be represented as:
\begin{equation}
\resizebox{\columnwidth}{!}{%
\(
\mathcal{L}_{\text{contrast}} = \lambda \left( \text{CrossEntropy}(\text{sim}, \mathbf{y}) + \text{CrossEntropy}(\text{sim}^\top, \mathbf{y}) \right)
\)
}.
\end{equation}

where the coefficient $\lambda$ is set to 0.5.

\begin{table*}[t]
\centering
\caption{
\textbf{Comparison on BABEL \cite{punnakkal2021babel}.} The right arrow $\rightarrow$ indicates that closer values to real motion are better. \textbf{Bold} and \underline{underline} highlight the best and second-best results, respectively. Additionally, $*$ denotes results reproduced by FlowMDM. SLI denotes spherical linear interpolation. For results with $\pm{0.000}$ or $\pm{0.00}$, the corresponding paper does not provide error bars.}
\label{tab:compare}
\resizebox{\textwidth}{!}{ %
\begin{tabular}{l|cccc|cccc}
\toprule
\multicolumn{1}{c}{} & \multicolumn{4}{c}{Subsequence} & \multicolumn{4}{c}{Transition} \\
Models  & R-precision $\uparrow{}$ & FID $\downarrow{}$ & Diversity $\rightarrow$ & MM-Dist $\downarrow{}$ & FID $\downarrow{}$ & Diversity $\rightarrow{}$ & PJ $\rightarrow{}$ & AUJ $\downarrow{}$ \\
\midrule
Ground Truth  & $0.715^{\pm0.003}$ & $0.00^{\pm0.00}$ & $8.42^{\pm0.15}$ & $3.36^{\pm0.00}$ & $0.00^{\pm0.00}$ & $6.20^{\pm0.06}$ & $0.02^{\pm0.00}$ & $0.00^{\pm0.00}$\\
\midrule
TEACH \cite{athanasiou2022teach} & $0.460^{\pm0.000}$ & $1.12^{\pm0.00}$ & $8.28^{\pm0.00}$ & $7.14^{\pm0.00}$ & $7.93^{\pm0.00}$ & $6.53^{\pm0.00}$ & -- & --\\
TEACH w/o SLI \cite{athanasiou2022teach} & $\mathbf{0.703}^{\pm0.002}$ & $1.71^{\pm0.03}$ & $8.18^{\pm0.14}$ & $\underline{3.43}^{\pm0.01}$ & $3.01^{\pm0.04}$ & $\mathbf{6.23}^{\pm0.05}$ & $1.09^{\pm0.00}$ & $2.35^{\pm0.01}$\\
TEACH$^*$ \cite{athanasiou2022teach} & $0.655^{\pm0.002}$ & $1.82^{\pm0.02}$ & $7.96^{\pm0.11}$ & $3.72^{\pm0.01}$ & $3.27^{\pm0.04}$ & $6.14^{\pm0.06}$ & ${0.07}^{\pm0.00}$ & ${0.44}^{\pm0.00}$ \\
PriorMDM \cite{shafir2023human} & $0.430^{\pm0.000}$ & $1.04^{\pm0.00}$ & $8.14^{\pm0.00}$ & $7.39^{\pm0.00}$ & $3.45^{\pm0.00}$ & $7.19^{\pm0.00}$ & -- & --\\
PriorMDM w/ Trans. Emb \cite{shafir2023human} & $0.480^{\pm0.000}$ & $0.79^{\pm0.00}$ & $8.16^{\pm0.00}$ & $6.97^{\pm0.00}$ & $7.23^{\pm0.00}$ & $6.41^{\pm0.00}$ & -- & --\\
PriorMDM w/ Trans. Emb \& geo losses \cite{shafir2023human} & $0.450^{\pm0.000}$ & $0.91^{\pm0.00}$ & $8.16^{\pm0.00}$ & $7.09^{\pm0.00}$ & $6.05^{\pm0.00}$ & $6.57^{\pm0.00}$ & -- & -- \\
PriorMDM$^*$ \cite{shafir2023human} & $0.596^{\pm0.005}$ & $3.16^{\pm0.06}$ & $7.53^{\pm0.11}$ & $4.17^{\pm0.02}$ & $3.33^{\pm0.06}$ & $\underline{6.16}^{\pm0.05}$ & $0.28^{\pm0.00}$ & $1.04^{\pm0.01}$\\
PriorMDM w/ PCCAT and APE \cite{shafir2023human} & $0.668^{\pm0.005}$ & $1.33^{\pm0.04}$ & $7.98^{\pm0.12}$ & $3.67^{\pm0.03}$ & $3.15^{\pm0.05}$ & $6.14^{\pm0.07}$ & $0.17^{\pm0.00}$ & $0.64^{\pm0.01}$\\
MultiDiffusion \cite{bar2023multidiffusion} & $\underline{0.702}^{\pm0.005}$ & $1.74^{\pm0.04}$ & $\underline{8.37}^{\pm0.13}$ & $\underline{3.43}^{\pm0.02}$ & $6.56^{\pm0.12}$ & $5.72^{\pm0.07}$ & $0.18^{\pm0.00}$ & $0.68^{\pm0.00}$\\
DiffCollage \cite{zhang2023diffcollage} & $0.671^{\pm0.003}$ & $1.45^{\pm0.05}$ & $7.93^{\pm0.09}$ & $3.71^{\pm0.01}$ & $4.36^{\pm0.09}$ & $6.09^{\pm0.08}$ & $0.19^{\pm0.00}$ & $0.84^{\pm0.01}$\\
T2LM \cite{lee2024t2lm} & $0.589^{\pm0.000}$ & $0.66^{\pm0.00}$ & $8.99^{\pm0.00}$ & $3.81^{\pm0.00}$ & -- & -- & -- & --\\
FlowMDM \cite{barquero2024seamless} & $\underline{0.702}^{\pm0.004}$ & $0.99^{\pm0.04}$ & $8.36^{\pm0.13}$ & $3.45^{\pm0.02}$ & $\underline{2.61}^{\pm0.06}$ & $6.47^{\pm0.05}$ & $\mathbf{0.06}^{\pm0.00}$ & $\underline{0.13}^{\pm0.00}$\\
Motion Mamba \cite{zhang2025motion} & $0.490^{\pm0.000}$ & $0.76^{\pm0.00}$ & $\mathbf{8.39}^{\pm0.00}$ & $4.97^{\pm0.00}$ & -- & -- & -- & --\\
InfiniMotion \cite{zhang2024infinimotion} & $0.510^{\pm0.000}$ & $\underline{0.58}^{\pm0.00}$ & $8.67^{\pm0.00}$ & $4.89^{\pm0.00}$ & -- & -- & -- & --\\
\midrule
\textbf{KMM} (Ours) & $0.666^{\pm0.001}$ & $\mathbf{0.34}^{\pm0.01}$ & $8.67^{\pm0.14}$ & $\mathbf{3.11}^{\pm0.01}$ & $\mathbf{1.37}^{\pm0.04}$ & ${5.96}^{\pm0.09}$ & $\underline{0.08}^{\pm0.00}$ & $\mathbf{0.10}^{\pm0.00}$\\
\bottomrule
\end{tabular}
}
\vskip -0.1in
\end{table*}

\section{Experiments}

\subsection{Datasets and Evaluation Matrices}

\paragraph{BABEL Dataset.} BABEL \cite{punnakkal2021babel} is the go-to benchmark for long motion generation and has been widely adopted in previous extended motion generation work. Derived from AMASS \cite{mahmood2019amass}, BABEL provides both frame-level and motion annotations for extended motion sequences. The dataset includes a total of 10,881 motion sequences, consisting of 65,926 segments, each with its corresponding textual label.

\paragraph{BABEL-D Dataset.} To evaluate the performance of text-motion alignment in extended motion generation methods, we introduce a new benchmark, BABEL-D. This benchmark is a subset of the BABEL test set and includes directional conditions with keywords such as \textit{left} and \textit{right}. This also represents the more challenging subset of BABEL. The BABEL-D dataset contains a total of 560 motion segments, enabling us to demonstrate improved alignment between generated motion and given text queries. We then evaluate our method's performance on BABEL-D and compare it with other state-of-the-art extended motion generation approaches.

\paragraph{HumanML3D Dataset.} HumanML3D \cite{guo2022generating} is the go-to dataset for text-to-motion generation, including 14,616 motions with text descriptions. Despite the maximum length of HumanML3D being only 196 frames, we also evaluate our method on this dataset to demonstrate its generalizability.

\paragraph{Evaluation Matrices.} For our experiments, we adopted the quantitative evaluation matrices for text-to-motion generation originally introduced by T2M \cite{guo2022generating} and later used in long motion generation studies \cite{shafir2023human,barquero2024seamless,zhang2024infinimotion}. These include: (1) Frechet Inception Distance (FID), which measures overall motion quality by assessing the distributional difference between the high-level features of generated and real motions; (2) R-precision; (3) MultiModal Distance, both of which evaluate the semantic alignment between the input text and generated motions; and (4) Diversity, which calculates the variance in features extracted from the motions. For transition evaluation, we adopt two metrics from FlowMDM \cite{barquero2024seamless}. Peak Jerk (PJ) captures the maximum jerk across joints to identify abrupt changes. However, it may favor overly smoothed transitions. To address this, we also include Area Under the Jerk (AUJ), which quantifies deviations from average jerk using L1-norm differences.

\subsection{Comparative Study}

\paragraph{Evaluation on BABEL.} To evaluate the performance of our KMM on extended motion generation, we trained and evaluated it on the BABEL dataset. The results, as shown in Tables \ref{tab:compare}, indicate that our method significantly outperforms previous text-to-motion generation approaches specifically designed for long-sequence motion generation. All experiments were conducted with a batch size of 256 for VQ-VAE, which utilized 6 quantization layers, and a batch size of 64 for 
mask bidirectional Mamba.
These experiments were carried out on a single Intel Xeon Platinum 8360Y CPU at 2.40GHz, paired with a single NVIDIA A100 40G GPU and 32GB of RAM.

\paragraph{Evaluation on BABEL-D.} To quantitatively demonstrate the advantages of our proposed text-motion alignment method in addressing directional instructions, we conducted comprehensive experiments on the newly introduced BABEL-D benchmark. The results have shown in Table \ref{tab:compare_d}. Compared to previous state-of-the-art methods, our approach significantly outperforms other extended motion generation techniques, indicating a stronger alignment between text and motion.

\begin{table}[H]
\caption{
\textbf{Comparison on BABEL-D.} The right arrow $\rightarrow$ indicates that closer values to real motion are better. \textbf{Bold} and \underline{underline} highlight the best and second-best results, respectively.}
\label{tab:compare_d}
\centering
\resizebox{\linewidth}{!}{ %
\begin{tabular}{l|cccc}
\toprule
Models  & R-precision $\uparrow{}$ & FID $\downarrow{}$ & Diversity $\rightarrow$ & MM-Dist $\downarrow{}$ \\
\midrule
Ground Truth  & $0.438^{\pm0.000}$ & $0.02^{\pm0.00}$ & $8.46^{\pm0.00}$ & $3.71^{\pm0.00}$ \\
\midrule
PriorMDM \cite{shafir2023human} & $0.334^{\pm0.015}$ & $6.82^{\pm0.76}$ & $7.27^{\pm0.33}$ & $7.44^{\pm0.12}$ \\
FlowMDM \cite{barquero2024seamless} & $\underline{0.535}^{\pm0.010}$ & $\underline{1.45}^{\pm0.07}$ & $\underline{8.09}^{\pm0.09}$ & $\underline{2.87}^{\pm0.03}$\\
KMM w/o Alignment & ${0.484}^{\pm0.007}$ & ${5.50}^{\pm0.15}$ & $\mathbf{8.44}^{\pm0.15}$ & ${3.48}^{\pm0.03}$\\
\midrule
\textbf{KMM} (Ours) & $\mathbf{0.538}^{\pm0.009}$ & $\mathbf{0.62}^{\pm0.03}$ & ${8.04}^{\pm0.14}$ & $\mathbf{2.72}^{\pm0.03}$\\
\bottomrule
\end{tabular}
}
\end{table}

\paragraph{Evaluation on HumanML3D.} We conducted experiments on HumanML3D \cite{guo2022generating} and compared our results with previous state-of-the-art long-motion methods. The results are presented in the Table \ref{tab:humanml}, indicating that our KMM method significantly outperforms previous long-motion methods and demonstrates strong generalizability across multiple datasets.

\begin{table}

\caption{\textbf{Comparison on HumanML3D \cite{guo2022generating}.} The right arrow $\rightarrow$ indicates that closer values to real motion are better. \textbf{Bold} and \underline{underline} highlight the best and second-best results, respectively.}
\label{tab:humanml}
\centering
\resizebox{\linewidth}{!}{ %
\begin{tabular}{l|cccc}
\toprule
Models  & R-precision $\uparrow{}$ & FID $\downarrow{}$ & Diversity $\rightarrow$ & MM-Dist $\downarrow{}$ \\
\midrule
Ground Truth  & $0.796^{\pm0.004}$ & $0.00^{\pm0.00}$ & $9.34^{\pm0.08}$ & $2.97^{\pm0.01}$ \\
\midrule
MultiDiffusion \cite{bar2023multidiffusion} & $0.629^{\pm0.002}$ & $1.19^{\pm0.03}$ & $\underline{9.38}^{\pm0.08}$ & $4.02^{\pm0.01}$ \\
DiffCollage \cite{zhang2023diffcollage} & $0.615^{\pm0.005}$ & $1.56^{\pm0.04}$ & $8.79^{\pm0.08}$ & $4.13^{\pm0.02}$ \\
PriorMDM \cite{shafir2023human} & $0.590^{\pm0.000}$ & $0.60^{\pm0.00}$ & $9.50^{\pm0.00}$ & $5.61^{\pm0.00}$ \\
FlowMDM \cite{barquero2024seamless} & $\underline{0.685}^{\pm0.004}$ & $\underline{0.29}^{\pm0.01}$ & ${9.58}^{\pm0.12}$ & $\underline{3.61}^{\pm0.01}$\\
\midrule
\textbf{KMM} (Ours) & $\mathbf{0.787}^{\pm0.005}$ & $\mathbf{0.15}^{\pm0.01}$ & $\mathbf{9.37}^{\pm0.01}$ & $\mathbf{3.08}^{\pm0.02}$\\
\bottomrule
\end{tabular}
}
\end{table}

\begin{figure*}[t]
    \centering
    \includegraphics[width=\linewidth]{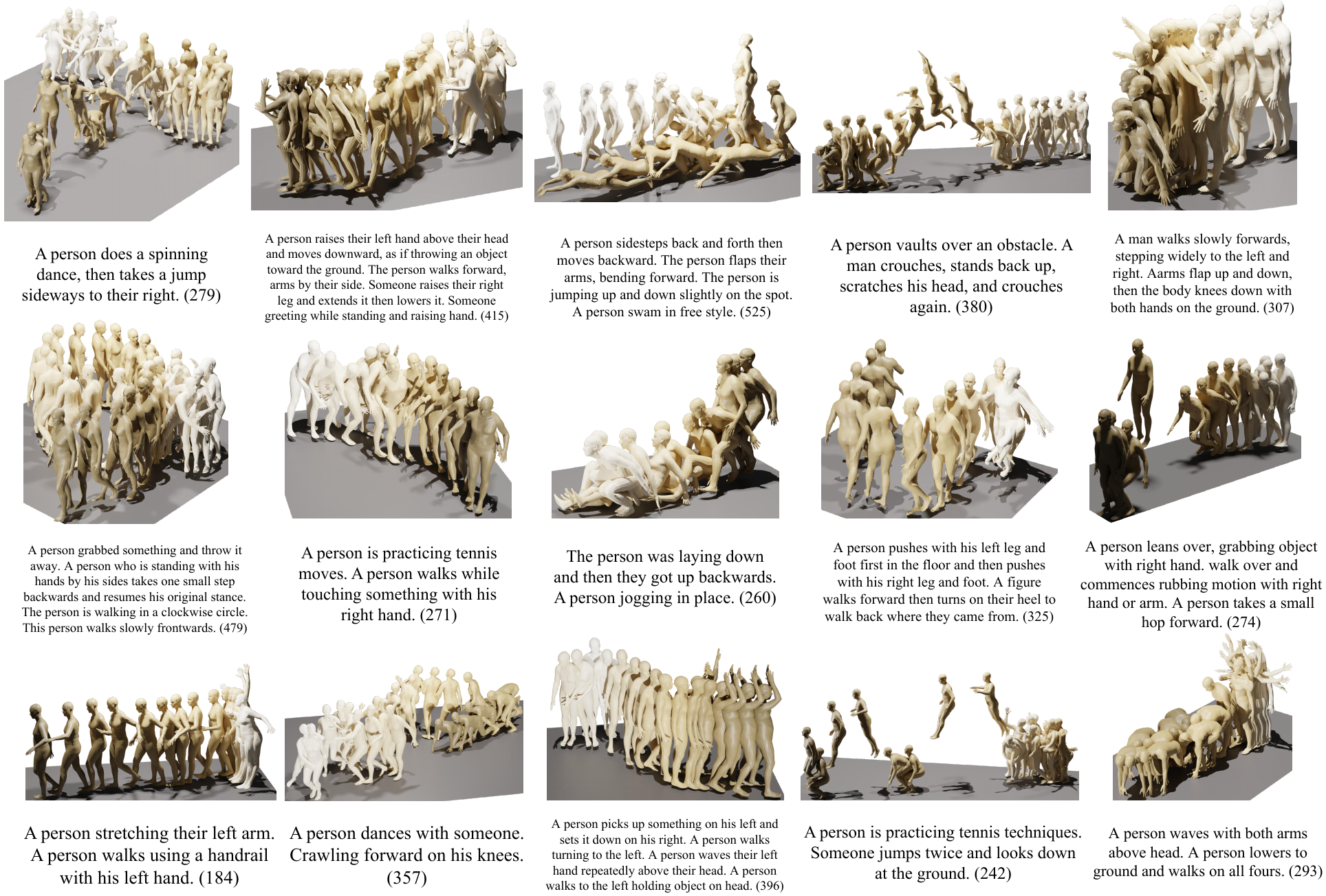}
    \caption{The figure presents some qualitative visualization results of KMM. The text prompts are sourced and combined from HumanML3D \cite{guo2022generating} and BABEL \cite{punnakkal2021babel}. The number within the brackets indicates our ability to condition the generated motion on a specific length, dynamically producing motion of the desired duration. The visualizations showcase KMM's superior performance in generating robust and diverse motions that align closely with lengthy and complex text queries.}
    \label{fig:demo}
\end{figure*}

\subsection{Ablation Study}

To further evaluate different aspects of our method's impact on overall performance, we conducted various ablation studies on the BABEL \cite{punnakkal2021babel}, as shown in Table \ref{tab:ablation}. The results show that our approach substantially outperforms other masking strategies, including random masking, KMeans \cite{lloyd1982least}, and GMM \cite{reynolds2009gaussian} key frame selection. Additionally, our proposed text-motion alignment framework greatly improves the model's ability to understand complex text queries, leading to better-aligned motion sequences.

\begin{table}
\caption{
\textbf{Masking strategies.} The right arrow $\rightarrow$ indicates that closer values to real motion are better. \textbf{Bold} and \underline{underline} highlight the best and second-best results, respectively.}
\label{tab:ablation}
\centering
\resizebox{\linewidth}{!}{ %
\begin{tabular}{l|cccc}
\toprule
Models  & R-precision $\uparrow{}$ & FID $\downarrow{}$ & Diversity $\rightarrow$ & MM-Dist $\downarrow{}$ \\
\midrule
Ground Truth  & $0.715^{\pm0.003}$ & $0.00^{\pm0.00}$ & $8.42^{\pm0.15}$ & $3.36^{\pm0.00}$ \\
\midrule
KMM w/ random masking  & $0.649^{\pm0.001}$ & $0.48^{\pm0.01}$ & $8.80^{\pm0.06}$ & $3.30^{\pm0.01}$ \\
KMM w/ KMeans & $\underline{0.661}^{\pm0.001}$ & $0.43^{\pm0.07}$ & $\mathbf{8.38}^{\pm0.09}$ & $3.16^{\pm0.01}$ \\
KMM w/ GMM  & ${0.659}^{\pm0.002}$ & $\underline{0.40}^{\pm0.01}$ & $\underline{8.30}^{\pm0.26}$ & $\underline{3.12}^{\pm0.01}$ \\
KMM w/o Alignment & $\underline{0.661}^{\pm0.001}$ & $\underline{0.40}^{\pm0.01}$ & ${8.57}^{\pm0.05}$ & ${3.21}^{\pm0.01}$\\
\midrule
\textbf{KMM} (Ours) & $\mathbf{0.666}^{\pm0.001}$ & $\mathbf{0.34}^{\pm0.01}$ & ${8.67}^{\pm0.14}$ & $\mathbf{3.11}^{\pm0.01}$\\
\bottomrule
\end{tabular}
}
\end{table}

We also conducted comprehensive ablation studies on the key frame masking ratio and the coefficient $\lambda$ in the contrastive loss for text-motion alignment on BABEL \cite{punnakkal2021babel}. The results are shown in Tables \ref{tab:maskratio} and \ref{tab:lambda}, demonstrating that our method is robust across different hyperparameter settings.

\section{Qualitative Evaluation}

\begin{table}
\caption{\textbf{Masking ratio.}
The right arrow $\rightarrow$ indicates that closer values to real motion are better. 
\textbf{Bold} highlights the best results.
}
\label{tab:maskratio}
\centering
\resizebox{\linewidth}{!}{ %
\begin{tabular}{l|cccc}
\toprule
Masking Ratio  & R-precision $\uparrow{}$ & FID $\downarrow{}$ & Diversity $\rightarrow$ & MM-Dist $\downarrow{}$ \\
\midrule
Ground Truth  & $0.715^{\pm0.003}$ & $0.00^{\pm0.00}$ & $8.42^{\pm0.15}$ & $3.36^{\pm0.00}$ \\
\midrule
15\%  & $0.661^{\pm0.001}$ & $0.69^{\pm0.01}$ & ${8.33}^{\pm0.15}$ & $3.27^{\pm0.01}$ \\
\rowcolor{yellow} $\mathbf{30\%}$ (Ours) & $\mathbf{0.666}^{\pm0.001}$ & $\mathbf{0.34}^{\pm0.01}$ & $\mathbf{8.67}^{\pm0.14}$ & $\mathbf{3.11}^{\pm0.01}$ \\
50\% & $0.063^{\pm0.003}$ & $0.41^{\pm0.01}$ & $8.79^{\pm0.01}$ & $3.26^{\pm0.01}$ \\
\bottomrule
\end{tabular}
}
\end{table}

\begin{table}
\caption{\textbf{Coefficient} $\lambda$.
The right arrow $\rightarrow$ indicates that closer values to real motion are better. 
\textbf{Bold} highlights the best results.
}
\label{tab:lambda}
\centering
\resizebox{\linewidth}{!}{ %
\begin{tabular}{l|cccc}
\toprule
Coefficient $\lambda$ & R-precision $\uparrow{}$ & FID $\downarrow{}$ & Diversity $\rightarrow$ & MM-Dist $\downarrow{}$ \\
\midrule
Ground Truth  & $0.715^{\pm0.003}$ & $0.00^{\pm0.00}$ & $8.42^{\pm0.15}$ & $3.36^{\pm0.00}$ \\
\midrule
0.3  & $0.667^{\pm0.003}$ & $0.40^{\pm0.01}$ & $8.64^{\pm0.08}$ & $3.25^{\pm0.01}$ \\
\rowcolor{yellow} $\mathbf{0.5}$ (Ours) & $\mathbf{0.666}^{\pm0.001}$ & $\mathbf{0.34}^{\pm0.01}$ & $\mathbf{8.67}^{\pm0.14}$ & $\mathbf{3.11}^{\pm0.01}$ \\
0.7 & $0.680^{\pm0.003}$ & $0.39^{\pm0.01}$ & $8.83^{\pm0.04}$ & $3.25^{\pm0.01}$ \\
\bottomrule
\end{tabular}
}
\end{table}

To further evaluate our method qualitatively, we compared KMM with TEACH \cite{athanasiou2022teach}, PriorMDM \cite{shafir2023human}, and FlowMDM \cite{barquero2024seamless} by generating a diverse set of prompts, randomly extracted and combined from the HumanML3D \cite{guo2022generating} and BABEL \cite{punnakkal2021babel} test sets. Figure \ref{fig:qualitative} shows three of these comparisons, demonstrating that our method significantly outperforms others in handling complex text queries and generating more accurate corresponding motions. Moreover, to further demonstrate the robustness and diversity of motions generated by our KMM, we produced 15 additional sequences using text prompts randomly extracted and combined from the HumanML3D \cite{guo2022generating} and BABEL \cite{punnakkal2021babel} test sets. The results, shown in figure \ref{fig:demo}, highlight superior performance in generating robust and diverse motions that closely align with lengthy and complex text queries.

\section{Conclusion}

In conclusion, our study addresses two significant challenges in extended motion generation: memory limitations of Mamba's hidden state for long sequence generation and weak text-motion alignment. Our proposed method, KMM, presents innovative solutions that significantly advance the field. Our density-based key frame selection and masking strategy enhances Mamba's ability to focus on critical actions within long motion sequences, effectively mitigating the memory limitation problem. Additionally, our robust contrastive learning paradigm improves text-motion alignment, enabling more accurate motion generation for complex and directional text queries. Furthermore, the development of the BABEL-D benchmark provides a valuable resource for evaluating text-motion alignment in extended motion generation, specifically focused on directional instructions. This new dataset, alongside our comprehensive experiments on the BABEL dataset, underscores our commitment to advancing the field of motion generation across various domains.


\appendix
\vspace{0.5cm}
{\LARGE\textbf{Appendix}}
\vspace{-0.3cm}

\begin{figure*}[t]
    \centering
    \includegraphics[width=\linewidth]{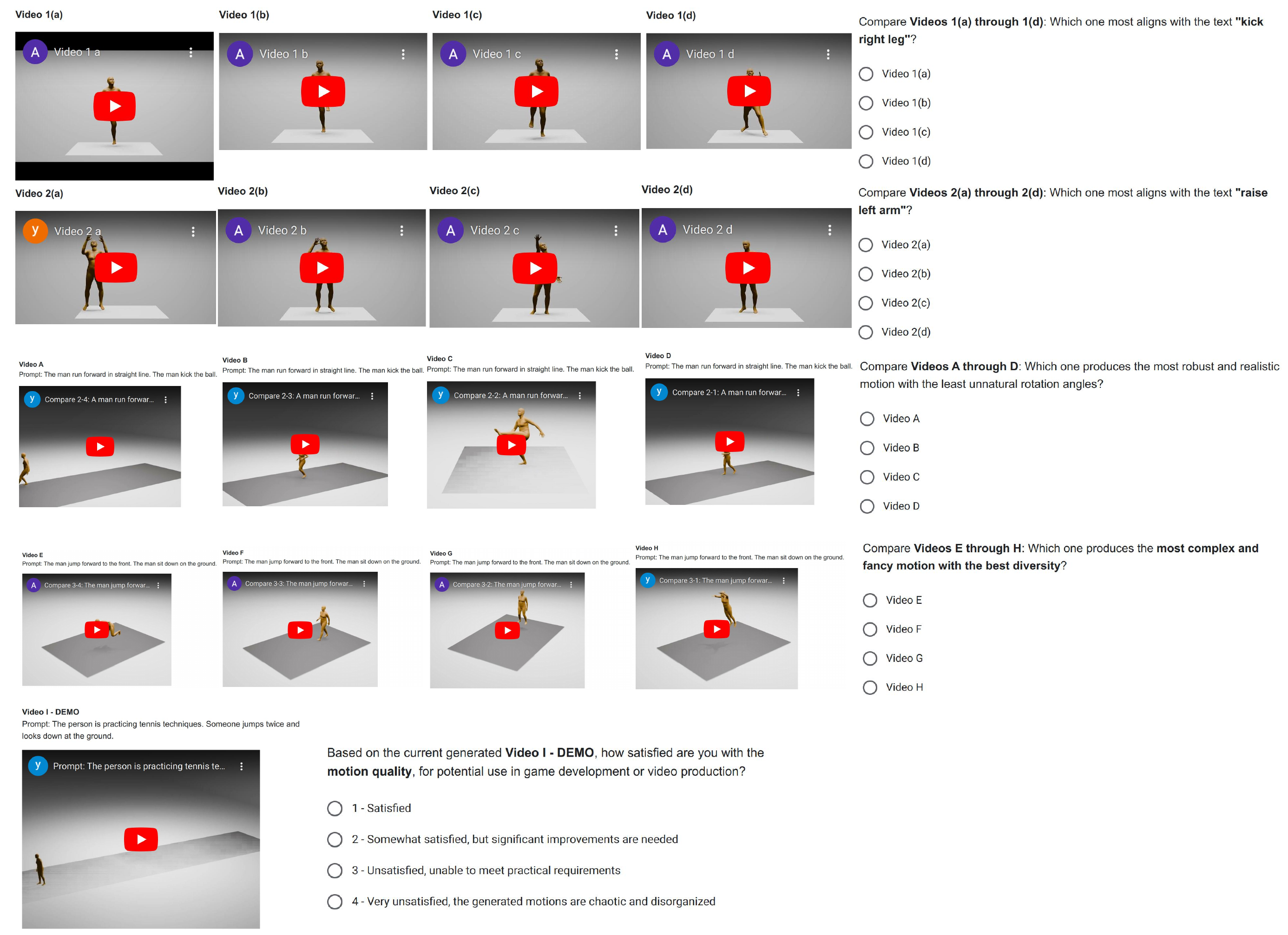}
    \caption{The figure shows the user study interface where 50 participants evaluated motion sequences generated by TEACH, PriorMDM, FlowMDM, and KMM, focusing on text-motion alignment, robustness, diversity, and usability. The text prompt are randomly extracted and combined from the HumanML3D \cite{guo2022generating} and BABEL \cite{punnakkal2021babel} test set.}
    \label{fig:user}
\end{figure*}

\section{User Study}

In this work, we conduct a comprehensive evaluation of KMM’s performance through both qualitative analyses across various datasets and a user study to assess its real-world applicability. We generated a diverse set of 15 motion sequences, randomly extracted and combined from the HumanML3D \cite{guo2022generating} and BABEL \cite{punnakkal2021babel} test set, using three different methods: TEACH \cite{athanasiou2022teach}, PriorMDM \cite{shafir2023human}, and FlowMDM \cite{barquero2024seamless}, alongside the generative results of KMM.

\textbf{Fifty participants} were randomly selected to evaluate the motion sequences generated by these methods. The user study was conducted via a Google Forms interface, as shown in figure \ref{fig:user}, ensuring that the sequences were presented anonymously without revealing their generative model origins. Our analysis centered on four key dimensions: 

\begin{itemize}
    \item The fidelity of text-motion alignment for directional instructions.
    \item The robustness of the generated motion.
    \item The diversity of the generated sequences.
    \item The overall performance and real-world usability.
\end{itemize}

\noindent The results shows that:
\begin{itemize}
    \item There is \textbf{92\%} of users who believe that KMM offers better motion alignment in directional instructions than other methods. 
    \item There is \textbf{78\%} of users who believe our method produces more robust and realistic motion with significantly fewer unrealistic rotation angles.
    \item There is \textbf{84\%} of users who believe that KMM generates more diverse and dynamic motion compared to the other three methods.
    \item For overall performance, there is \textbf{64\%} of users who believe that our generation results are satisfactory and have strong potential for real-world applications.
\end{itemize}

\end{document}